\documentclass[letterpaper, 10 pt, conference]{ieeeconf}  % Comment this line out if you need a4paper

\IEEEoverridecommandlockouts                              % This command is only needed if 
\usepackage{graphics} % for pdf, bitmapped graphics files
\usepackage{epsfig} % for postscript graphics files
\usepackage{mathptmx} % assumes new font selection scheme installed
\usepackage{times} % assumes new font selection scheme installed
\usepackage{amsmath} % assumes amsmath package installed
\usepackage{amssymb}  % assumes amsmath package installed
\usepackage{siunitx}
\usepackage{cite}

\title{\LARGE \bf
AvoidBench: A high-fidelity vision-based obstacle avoidance benchmarking suite for multi-rotors
}

\author{Hang Yu, Guido C. H. E de Croon and Christophe De Wagter%
\thanks{All authors are with Faculty of Aerospace Engineering, Delft University of Technology, 2629 HS Delft, The Netherlands. (email: {\tt\small h.y.yu@tudelft.nl};  {\tt\small g.c.h.e.deCroon@tudelft.nl}; {\tt\small c.dewagter@tudelft.nl}).}%
}

\begin{document}

\maketitle
\thispagestyle{empty}
\pagestyle{empty}

%%%%%%%%%%%%%%%%%%%%%%%%%%%%%%%%%%%%%%%%%%%%%%%%%%%%%%%%%%%%%%%%%%%%%%%%%%%%%%%%
\begin{abstract}

Obstacle avoidance is an essential topic in the field of autonomous drone research. When choosing an avoidance algorithm, many different options are available, each with their advantages and disadvantages. As there is currently no consensus on testing methods, it is quite challenging to compare the performance between algorithms. In this paper, we propose AvoidBench, a benchmarking suite which can evaluate the performance of vision-based obstacle avoidance algorithms by subjecting them to a series of tasks. Thanks to the high fidelity of multi-rotors dynamics from RotorS and virtual scenes of Unity3D, AvoidBench can realize realistic simulated flight experiments. Compared to current drone simulators, we propose and implement both performance and environment metrics to reveal the suitability of obstacle avoidance algorithms for environments of different complexity. To illustrate AvoidBench's usage, we compare three algorithms: Ego-planner, MBPlanner, and Agile-autonomy. The trends observed are validated with real-world obstacle avoidance experiments.

Code: https://github.com/tudelft/AvoidBench
\end{abstract}

%%%%%%%%%%%%%%%%%%%%%%%%%%%%%%%%%%%%%%%%%%%%%%%%%%%%%%%%%%%%%%%%%%%%%%%%%%%%%%%%
\section{INTRODUCTION}
Autonomous drones can be applied in several novel fields such as forest rescue, cave exploration \cite{dang2020graph, kulkarni2022autonomous, dharmadhikari2020motion}, and greenhouse or warehouse monitoring \cite{mlambo2017structure}. One common theme between all these cases is that obstacle avoidance plays a large role for the safety and effectivity of these vehicles. In particular, vision-based obstacle avoidance is popular, since vision sensor can be compact, light-weight and low cost. Unfortunately, there is currently no shared consensus on how vision-based obstacle avoidance algorithms should be tested. This makes it not only difficult to see how the field of obstacle avoidance as a whole is progressing, but it also makes it hard to compare the performance of different obstacle avoidance algorithms. There is a dire need for reliable benchmarking.

There are many benchmarks for fields like computer vision or natural language processing (NLP), such as KITTi \cite{geiger2012we}, ImageNet \cite{deng2009imagenet} and GLUE \cite{wang2018glue}. They all provide large datasets and proper metrics for evaluating and comparing different algorithms. This allows researchers not only to compare state-of-the-art algorithms but also to see how performance has increased over time. Furthermore, it offers an invaluable platform for researchers to showcase and share their algorithms and garner more interest from the community. If executed well, a benchmark can give a boost to the entire community and research field.

   \begin{figure}[thpb]
      \centering
      {\includegraphics[width=3.2in]{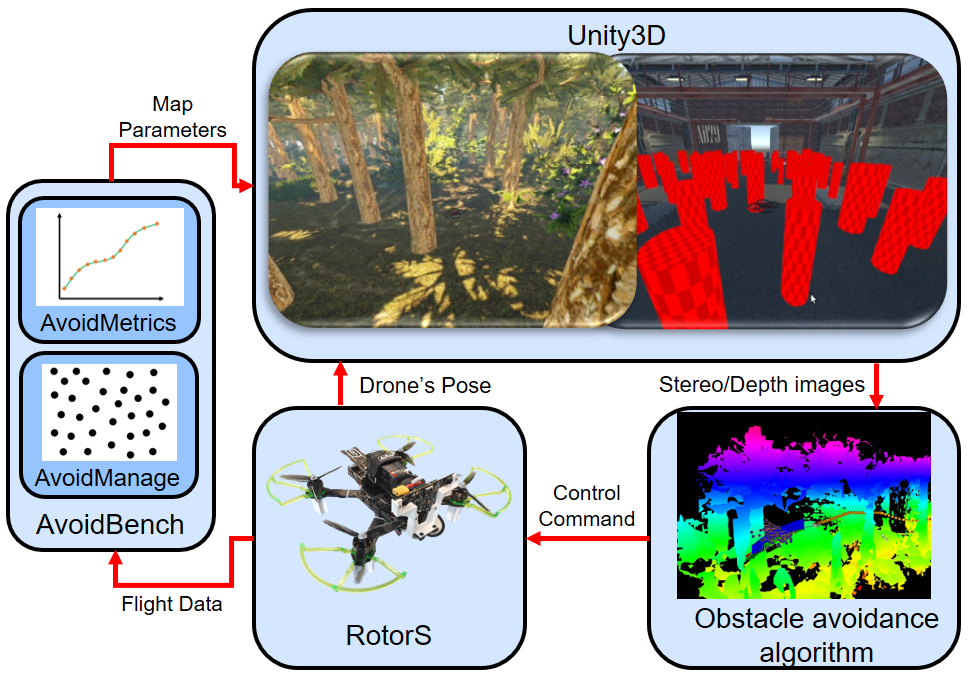}}
      \caption{The basic framework of AvoidBench. It uses the dynamics model from RotorS and photo-realistic scenes from Unity3D. Users can implement their obstacle avoidance algorithms by Python or C++ interfaces.}
      \label{figure1}
   \end{figure}
   
However, benchmarking algorithms for robotics is more difficult than for fields such as computer vision and NLP. This is mainly due to two reasons: (1) We cannot rely on a fixed dataset, since robots interact with their environment. Consequently, the sensory data and situations they encounter depend on the proposed autonomy algorithms. (2) A robot's success is dependent on a myriad of factors. These factors include real-world environment factors, ranging from its visual appearance to disturbances such as wind gusts, to an intricate internal pipeline with processes such as perception, planning and control. Hence, the number of interfering factors of performance are more than fields of computer vision and NLP.

There are already some mature simulators both for drones and rendering systems. For dynamics models, Gazebo-based simulators Hector \cite{meyer2012comprehensive} and RotorS \cite{furrer2016rotors} are both good choices, but Gazebo does not provide a photo-realistic visual environment. This low fidelity would lead to a large sim2real gap, and hence reduce the generalization of the benchmarking performances to real robotic platforms and environments.

The progress in the realism of game simulators is an opportunity for benchmarking in robotics. Specifically, rendering engines such as Unity3D and Unreal Engine generate photo-realistic environments. Airsim is a robot simulation platform developed by Microsoft and uses Unreal Engine 4 \cite{shah2018airsim}, and it provides a pure C++ and Python API to interface with the simulator. Another advantage of Airsim is that it supports hardware-in-the-loop (HITL) as well as software-in-the-loop (SITL) with flight controllers such as PX4. However, Airsim is tightly coupled with the rendering engine for the drone's dynamics and control system, so the simulation speed is limited and researchers usually use two devices to handle the rendering process and control process relatively. Flightmare \cite{song2020flightmare} is a more flexible simulator for quadrotors. It uses Unity3D for rendering and a Gazebo-based quadrotor dynamics \cite{furrer2016rotors} as well as a parallelized implementation of classical quadrotor dynamics model used for reinforcement learning.

There are also some benchmarks for autonomous drones. MAVBench \cite{boroujerdian2018mavbench} is a benchmark simulator to measure the performance of drones executing different missions such as search-and-rescue, mapping and planning. Using Airsim and Unreal Engine 4 as a backend, MAVBench is the first end-to-end photo-realistic benchmarking suite for MAVs. Although it supports many tasks, only time and energy are considered as the metrics and the complexity of environments is ignored. BOARR \cite{du2019boarr} is a benchmarking suite specifically designed to evaluate the performance of obstacle avoidance algorithms. Unfortunately, as BOARR is using Gazebo as a simulator, the visual fidelity is quite poor. We believe that in applications such as vision-based obstacle avoidance, visual realism plays a large role in how an algorithm performs.

In this article, we propose AvoidBench as a benchmarking suite for evaluating the performance of vision-based obstacle avoidance algorithms. We choose Flightmare as the basic backbone of AvoidBench, because it is lighter and can achieve higher simulation speed than Airsim. Based on Flightmare, we have for now built two simulation scenes for benchmarking: a forest environment and an indoor environment. It is easy to change the distribution of obstacles and complexity of map so that researchers can reveal the potential of drones using their algorithms. And we propose a complete set of metrics which contain the flight performance and environment complexity to evaluate the obstacle avoidance algorithms.

Basically, AvoidBench has the following three contributions: (1) We have developed a benchmark with high-fidelity visual scenes that allows to easily and objectively compare different algorithms in different conditions. The benchmark has been validated with real-world experiments; (2) Expanding upon Nous et al.'s work \cite{Masterthesis}, a complete set of evaluation metrics including the flight performance and environment complexity is proposed; (3) Different types of obstacle avoidance algorithms including learning-based, optimization-based and motion-primitive-based are tested in AvoidBench and the performance of each algorithm is evaluated objectively both in simulation and the real world; (4) We provide an open source software framework in which users can easily enter their algorithms in Python or C++, determining their performance in relation to various environment metrics.

\section{BENCHMARKING PIPELINE}

\begin{figure}[thpb]
      \centering
      {\includegraphics[width=2.8in]{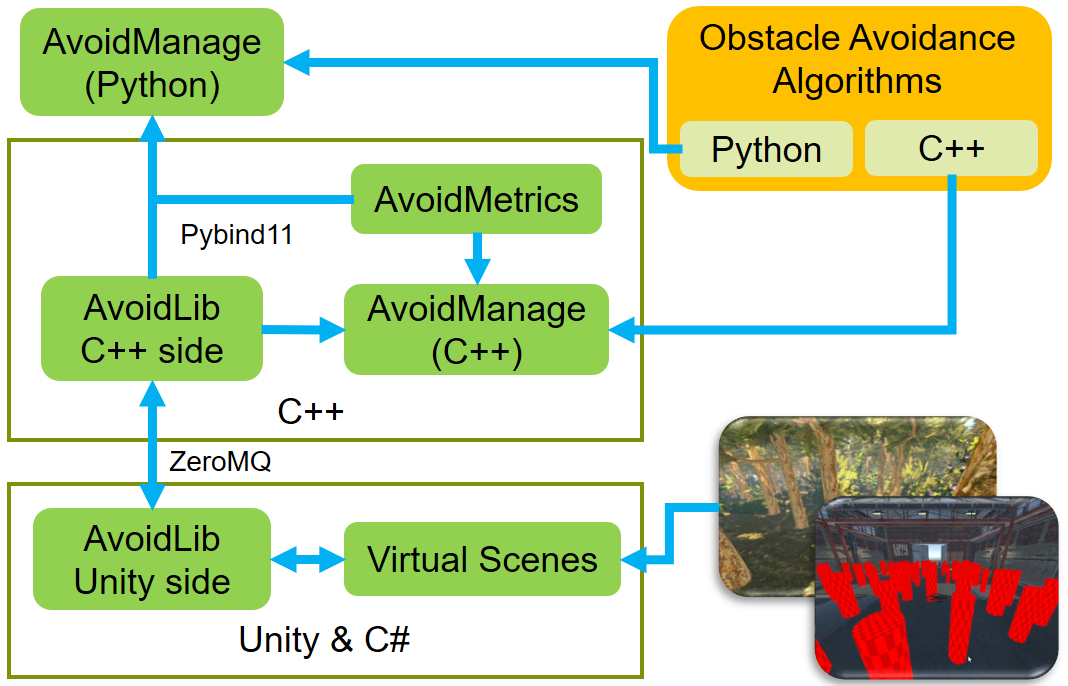}}
      \caption{Different modules of AvoidBench. AvoidLib is used to communicated between the Unity side and C++ side. AvoidManage contains the setup of virtual scenes and the whole process of flight and benchmarking. AvoidMetrics is the package of performance and environment metrics.}
      \label{figure2}
   \end{figure}

Based on Flightmare, we propose AvoidBench which can evaluate the performance of vision-based obstacle avoidance algorithms for multi-rotors in simulation. AvoidBench assigns performance scores to obstacle avoidance algorithms by subjecting them to a series of tasks. Due to the photo-realistic of Unity3D, we can minimize the discrepancy between simulated and real-world sensors’ observations. As shown in Figure \ref{figure2}, AvoidBench is a software suite that consists of four main modules: (1) virtual scenes based on Unity3D; (2) communication interface (AvoidLib) between the C++ side and the Unity side which is based on ZeroMQ; (3) AvoidManage which is used to combine the rendering engine and dynamics model as well as the management of the whole obstacle avoidance task; (4) AvoidMetrics which contains both performance and environment metrics to evaluate the obstacle avoidance task. In the remainder of this section, we explain each of the AvoidBench modules.

\subsection{Virtual Scenes}

The scene in AvoidBench is defined as an editable environment in which the obstacle avoidance algorithms are benchmarked. Due to the limited interface with the environment in Flightmare, it is difficult to meet our benchmarking requirements, such as adding terrain or bushes in the forest and changing the texture of obstacles in the indoor scene. Due to time restrictions, AvoidBench for now has two different environments: an outdoor scene and an indoor scene. As shown in Figure \ref{figure1}, the outdoor scene is a forest with the size of \SI{160}{m}$\times$\SI{160}{m} which is composed of red trees, bush trees and terrain, while the indoor scene is a warehouse with geometrical but differently textured obstacles. Here are some key concepts related to the setting of environments.

\subsubsection{Map}
The benefit of a simulator is that it is easy to create and auto-generate different maps for the flight tests. Users can easily change the number or position of outdoor trees as well as the size and opacity of texture on indoor obstacles. To evaluate the obstacle avoidance ability of drones, it is necessary to follow certain rules to generate obstacles. To generate the actual obstacle field, AvoidBench utilizes a rudimentary and well-known method for procedural generation named Poisson Disc Sampling \cite{bridson2007fast}. This method has a considerable advantage over random sampling as it creates a uniform distribution of obstacles. Furthermore, it allows us to control the minimum distance between obstacles.

   \begin{figure}[thpb]
      \centering
      {\includegraphics[width=3.2in]{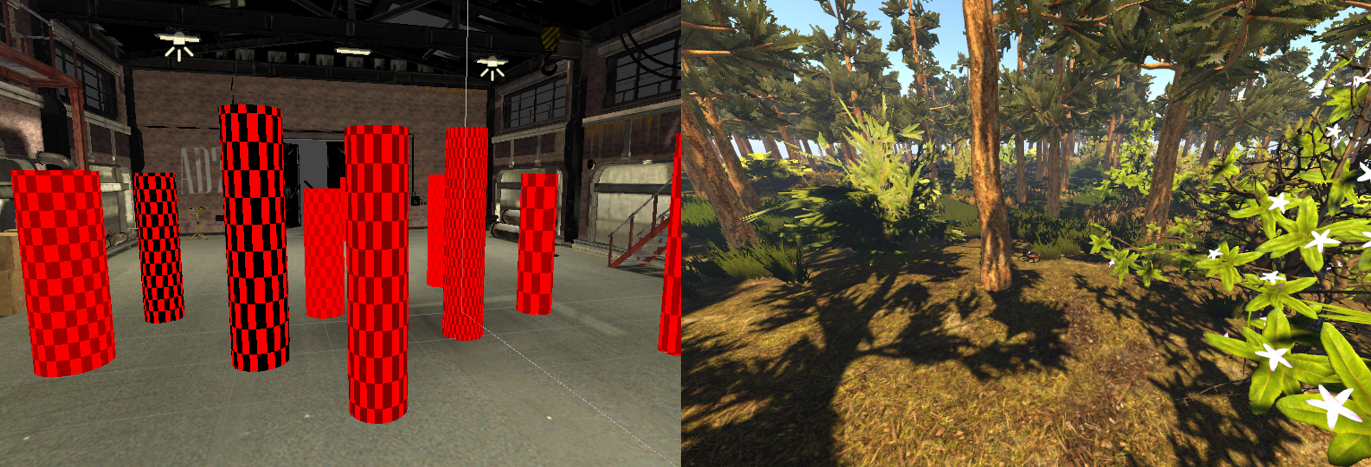}}
      \caption{Indoor and outdoor scenes. There are many geometrical obstacles but with different opacity of texture in indoor scene (left). Outdoor scene contains red trees and non-geometrically-shaped obstacles such as the bushes (right).}
      \label{figure3}
   \end{figure}
   
As shown in Figure \ref{figure3}, for the indoor scene, we use cylinders as the obstacles. Users can change the size and opacity of the texture on each cylinder as well as the radius of Poisson distribution. To get a new map we just need to set a new random seed. For the outdoor scene, we want to test the effectiveness of algorithms for nongeometrical obstacles, we set both red trees and bushes on the undulating terrain, users can also set the ratio of red trees to bush trees.

\subsubsection{Trial}
Given a general map, we perform different \emph{trials}. Each trial has a different start point and goal point generated by a random seed. The drone can traverse the whole map as much as possible. We use the mesh collider from Unity3D to check if the drone collided with obstacles. If the algorithm can command the drone to the goal without collision, this trial is marked as "\emph{finished}". Apart from collisions, there can be a variety of situations that lead to trial failure. An algorithm e.g., might get stuck in a dead end, unable to find or move to the goal. By setting a maximum time parameter, AvoidBench automatically stops the trial if the time limit has been exceeded.
 
\subsection{AvoidLib}

Like Flightmare, we also use ZeroMQ to build the communication interface between Unity and C++. ZeroMQ is an asynchronous messaging library, aimed at use in distributed or concurrent applications. Through AvoidLib, we can transfer the parameters of the map to Unity, update the pose of the drone in rendering engine, obtain the images collected by stereo cameras and get the collision detection results from Unity3D. Since the environment metrics which will be introduced below require an accurate point cloud map, we retain the function of saving point clouds in Flightmare.

\subsection{AvoidManage}

AvoidManage is used to combine the rendering engine with a simulated dynamics model of multi-rotors. As the management of whole benchmark, AvoidManage is also the state machine of the simulated drone. Firstly, it needs to get the control commands from an obstacle avoidance algorithm and obtain the flight data in real time. Then the flight data will be sent to AvoidMetrics to evaluate the performance of each \emph{trial} and determine if the \emph{trial} should be terminated according to the feedback of collision checking. Furthermore, we also use Pybind11 to wrap the AvoidLib and provide a Python version of AvoidManage for researchers that want to program their obstacle avoidance algorithms in Python.

\subsection{AvoidMetrics}

AvoidBench aims to capture the relation between avoidance performance and the environmental conditions. By measuring both performance and environment metrics, it becomes possible to relate the performance of an algorithm to the type of environment and its corresponding difficulty. The grand goal here is that based on extensive benchmarking results in simulation, we will be able to accurately predict an algorithm's performance in real-world environments.

\subsubsection{Environment Metrics}
AvoidBench currently measures two different environment metrics, named \emph{Traversability} and \emph{Relative Gap Size}.

\paragraph{Traversability}
\emph{Traversability} was introduced in \cite{nous2016performance} as a metric to quantify how difficult it would be for a drone to traverse an environment. It was presented as an alternative to obstacle density which had the issue that it did not accurately capture avoidance difficulty as the dimensions of the drone were not taken into account. Moreover, traversability does not depend on a specific obstacle geometry. It can be calculated in generic environments by picking $N$ random position $p$ in the environment. For each position, a random heading $h$ would be selected, and a collision checker is used to calculate the free-flight distance $s$. All the results are then averaged to obtain the uncorrected traversability. The final step to make the metric dimensionless was to divide the result by the drone diameter $d_{drone}$.

\begin{equation}
    TRAV = {\frac{1}{d_{drone} \cdot N}} \sum_{i=0}^N {s(p(i), h(i))}
    \label{eq6}
\end{equation}

In AvoidBench, we apply some changes to the original traversability metric to make the results more repeatable. Intuitively, the traversability represents how far the drone can travel without an active obstacle avoidance capability. Specifically, we use a grid-based sampling method instead of a uniformly random to sample locations. Additionally, at each location, instead of randomly picking a single direction vectors are generated evenly across a 2D circle. For each direction vector, the free-path length is calculated, and all results are averaged to obtain the traversability. Traversability is based on sampling trajectories, so that it can also be easily applied to more complex environments than those with cylindrical obstacles.

\paragraph{Relative Gap Size}
The second environment metric, \emph{relative gap size}, is less generic, as it is directly related to the radius of Poisson Distribution. This value represents the minimum space between obstacles expressed in multiples of the drone's diameter. It adjusts Poisson radius $r_{poisson}$ by subtracting the average width $\hat{w}_{obstacle}$ and divides it by the drone diameter $d_{drone}$:
\begin{equation}
    RGS = {\frac{1}{d_{drone} \cdot N}} \sum_{i=0}^N {s(p(i), h(i))}
    \label{eq7}
\end{equation}

The relative gap size should never be below a value of 1. In that case, the drone would be unable to traverse through the obstacle field as the obstacles are positioned too close together.

\subsubsection{Performance Metrics}
AvoidBench currently measures seven different types of performance metrics, which will be introduced as following.

\paragraph{Success Rate}

This metric is the most straightforward result of AvoidBench. Each map has several trials, according to the ratio of the number of "\emph{finished}" trials $T_{finished}$ to the whole number of trials $T_{whole}$, we can see the basic performance of this algorithm in the current map for a pre-selected start-to-target distance. The \emph{success rate} can be calculated as:

\begin{equation}
    SR = {\frac{T_{finished}}{T_{whole}}}
    \label{eq1}
\end{equation}

\paragraph{Path Optimality}
The \emph{path optimality} measures how optimal the drone's path is from the start point to the goal point. It can be calculated using:

\begin{equation}
    PO = {\frac{d_{trav}-d_{min}}{d_{min}}\times\SI{100}{\percent}}
    \label{eq2}
\end{equation}
Where $d_{trav}$ means the total distance travelled by the drone and $d_{min}$ is the minimum free-path distance from start point to end point calculated by a path searching algorithm (A* in our case). The path optimality should be interpreted as the excess distance covered by the drone relative to the shortest path. So, \SI{0}{\percent} means no excess distance, whereas a \SI{100}{\percent} means that the drone covered twice the distance it could have.

\paragraph{Energy Optimality}
The \emph{energy optimality} is used to measure the energy cost of drones when obstacles. We use the integral of jerk (derivation of acceleration) to express it. In real world, the lower of energy optimality means the longer drones can fly.

\begin{equation}
    EO = {\int_{0}^{t_{trial}}\Vert jerk \Vert_2\, \mathrm{d}t}
    \label{eq23}
\end{equation}

\paragraph{Average Goal Velocity}
The purpose of the \emph{average goal velocity} metric is to measure the speed with which an algorithm is able to traverse the obstacle field towards the goal point. Usually, the trial time it takes for the drone to travel from the start point to the goal point is used to evaluate the speed performance. However, to ensure that results are also comparable between different maps and path lengths, the trial time is converted to a velocity value using the free-path minimum distance, as shown in (\ref{eq2}). Here, $t_{trial}$ is defined as the trial time and the average goal velocity can be calculated as:

\begin{equation}
    AGV = {\frac{d_{min}}{t_{trial}}}
    \label{eq3}
\end{equation}

\begin{figure}[thpb]
      \centering
      {\includegraphics[width=1.4in]{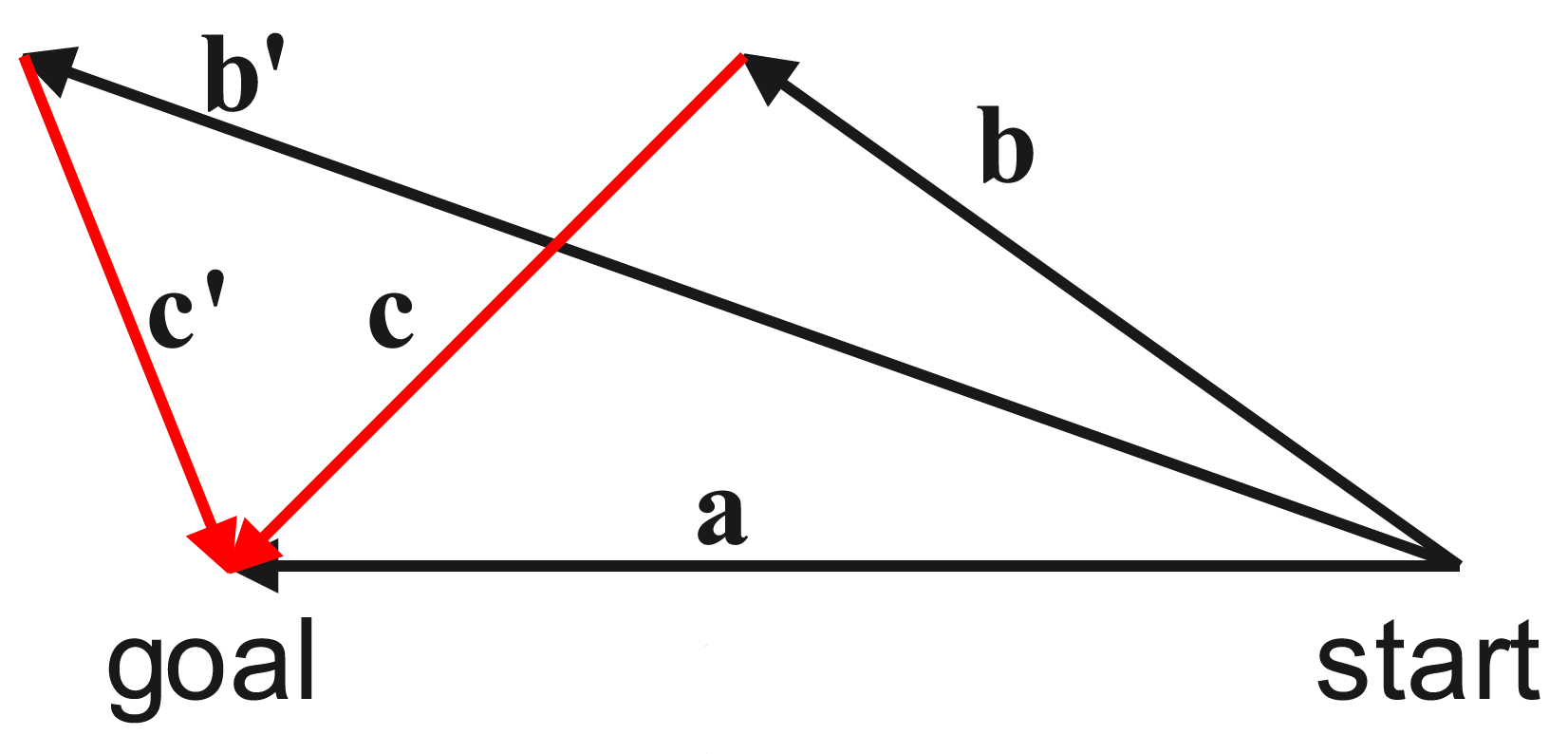}}
      \caption{Calculation method of mission progress. $\mathbf{a}$ is the vector of start point to goal point. $\mathbf{b}$ and $\mathbf{b^{\prime}}$ are both the vectors of start point to drone's final location.}
      \label{figure4}
   \end{figure}

\paragraph{Mission Progress}
The \emph{mission progress} metric measures how far the drone has progressed to the goal point, which is most useful for comparing runs when the drone does not reach it. In AvoidBench, \emph{mission progress} is calculated based on vector projection. Given vector $\mathbf{a}$ as the start-goal vector, $\mathbf{b}$ as the vector from start point to the drone's final location and $\mathbf{c} = \mathbf{a}-\mathbf{b}$ which means the vector from drone's final location to goal point shown as red lines in Figure \ref{figure4} (sometimes the drone stops behind the goal point, then use $\mathbf{b^{\prime}}$ and $\mathbf{c^{\prime}}$ to replace $\mathbf{b}$ and $\mathbf{c}$). The mission progress can be calculated by:

\begin{equation}
    MP = {(1-\lvert \frac{\mathbf{a}\cdot\mathbf{c}}{{\lvert \mathbf{a} \rvert}^2} \rvert)\times\SI{100}{\percent}}
    \label{eq5}
\end{equation}

\paragraph{Processing Time}
The last performance metric, \emph{processing time}, directly measures how long a single iteration of the obstacle avoidance algorithm takes. During the mission, all processing periods are stored, which allows AvoidBench to calculate statistics such as the mean and standard deviation of these values. The processing time is machine-specific, and hence algorithms would have to run on the same machine for a fair comparison. Processing time influences the control delay and as such already influences other performance metrics, such as success rate. However, it is also of interest by itself, as a lower processing time means that more computational resources are available for other autonomy tasks.

\paragraph{Contrast Factor}
The \emph{contrast factor} is the only metric that compares two algorithms. It can not only be used to measure the performance ratio of two different algorithms without the influence of flying distance, but also the performance differences of the same obstacle avoidance algorithm in the simulator and real world.
 
 From the perspective of probability, the success rate is related to the tested algorithms, the minimum free-path distance from start point to goal point, and the complexity of maps which can be represented by traversability. We use $\Xi$ to represent the obstacle avoidance algorithm and $\tau$ as a certain map. The normalized traversability $P_{\tau}$ can be calculated as:

\begin{equation}
    P_{\tau} = {\frac{TRAV}{TRAV_{max}}}
    \label{eq12}
\end{equation}
Where $TRAV_{max}$ is the value of traversaility when there is no obstacle in the map (only boundaries of the map have an impact on this value). Assuming the obstacles in the map are uniform, the probability of no collision for a certain algorithm within a certain distance is $p$, and the probability of no collision at twice distance should be $p^2$. Obviously, the success rate is exponentially related to the flying distance in theory. Then the success rate $\hat{SR}$ can be estimated as:

\begin{equation}
    \hat{SR}_{\Xi} = f\left(\Xi, P_{\tau}\right)^{\frac{^{\Xi}d_{min}}{\lambda}}
    \label{eq13}
\end{equation}
Where $\lambda$ is a scale factor of distance, and $^{\Xi}d_{min}$ is the minimum free-path distance from start point to goal point randomly generated for algorithm $\Xi$. $f\left(\Xi, P_{\tau}\right)$ indicates the non-collision probability for a certain distance determined by $\lambda$ when algorithm $\Xi$ is excuted in the map with normalized traversability $P_{\tau}$. From AvoidBench, $^{\Xi}d_{min}$ and $P_{\tau}$ are known, and $\hat{SR_{\Xi}}$ can be replaced by statistical results from (\ref{eq1}). However, we still cannot get $f\left(\Xi, P_{\tau}\right)$ because the $\lambda$ is unknown. By logarithmizing both sides of (\ref{eq13}), we can obtain the ratio of different algorithms' performance in the map with same traversability:

\begin{equation}
    CF = \frac{\ln{f\left(\Xi_{1}, P_{\tau}\right)}}{\ln{f\left(\Xi_{2}, P_{\tau}\right)}} = \frac{^{\Xi_{2}}d_{min}\ln{SR_{\Xi_{1}}}}{^{\Xi_{1}}d_{min}\ln{SR_{\Xi_{2}}}}
    \label{14}
\end{equation}

If $\Xi_{1}$ and $\Xi_{2}$ represent the experiments of the same algorithm in the simulator and real world, then contrast factor can be used to measure the sim2real gap even though they have different flying distance.

\section{SIMULATION AND EXPERIMENT}

\subsection{Obstacle Avoidance Algorithms}
In order to fully verify the configuration effect of obstacle avoidance algorithms on AvoidBench, we choose several mainstream, state-of-the-art obstacle avoidance algorithms, including: Agile-Autonomy \cite{Loquercio2021Science} (learning-based), Ego-planner \cite{zhou2020ego} (optimization-based), and MBPlanner \cite{dharmadhikari2020motion} (motion-primitive-based). For Agile-Autonomy, we use their original code and retrain the network through our scenes. MBPlanner is a exploration algorithm, we changed the exploration gain to the distance of current point to the goal point so that the drone can always fly to a fixed goal. All these three algorithms are tested by a virtual stereo camera which can generate depth map through the Semi-Global Matching algorithm \cite{hirschmuller2007stereo}.

   \begin{figure*}[thpb]
      \centering
        \includegraphics[width=6.8in]{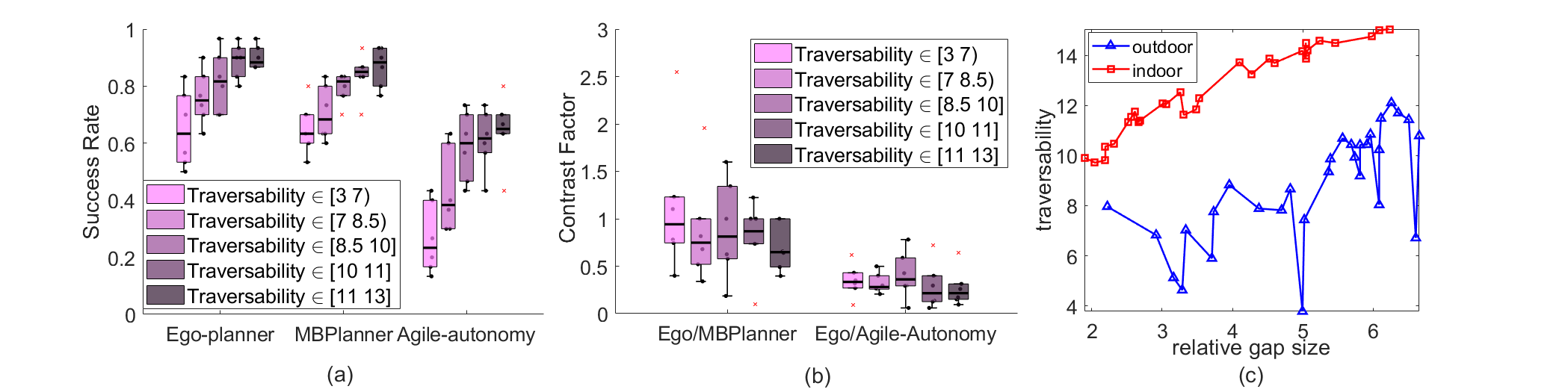}
    \caption{(a) is the success rate of different algorithms. (b) is the contrast factors of Ego-planner to MBPlanner and Ego-planner to Agile-Autonomy. (c) is the relationship between traversability and relative gap size of both indoor scene and outdoor scene.}
      \label{figure6}
   \end{figure*}
   
\subsection{Simulation Results}
   
For the outdoor scene, we set 30 maps and 30 trials. The ratio of bush trees to red trees is 2:3. The radius of the Poisson distribution is within the interval [2.3, 5.8]. Start points and end points are all randomly selected according to a random seed so that these three algorithms will encounter the same environments and tasks. All simulation experiments are run on a standard laptop with 16GB memory, i7 11800H CPU, and RTX3050 GPU.

Figure \ref{figure6}(a) shows the distribution of the success rate of the three algorithms with the increase of traversability. we divide the data into five groups so that it is easy to see the tendency. Obviously, for the three algorithms, success rate increases significantly with the increase of traversability, which shows that the performance of obstacle avoidance algorithms has a great relationship with the environment metrics we proposed. Figure \ref{figure6}(a) indicates that Ego-planner and MBPlanner have a similar performance concerning safety (success rate) and that both are better than Agile-Autonomy. Figure \ref{figure6}(b) is the result of contrast factor calculated by success rate. The value of "Ego / MBPlanner" is closed to 1, meaning that they have a similar performance while "Ego / Agile-Autonomy" is below 1, meaning that Ego-planner has a higher success rate than Agile-Autonomy. The relationship between traversability and relative gap size is shown in Figure \ref{figure6}(c). The traversability of outdoor scene does not always increase linearly according to relative gap size as the indoor scene because of nongeometrical bush trees. Basically, traversability is a more general metric to evaluate the complexity of environments.
   
\begin{figure*}[thpb]
      \centering
        \includegraphics[width=6.6in]{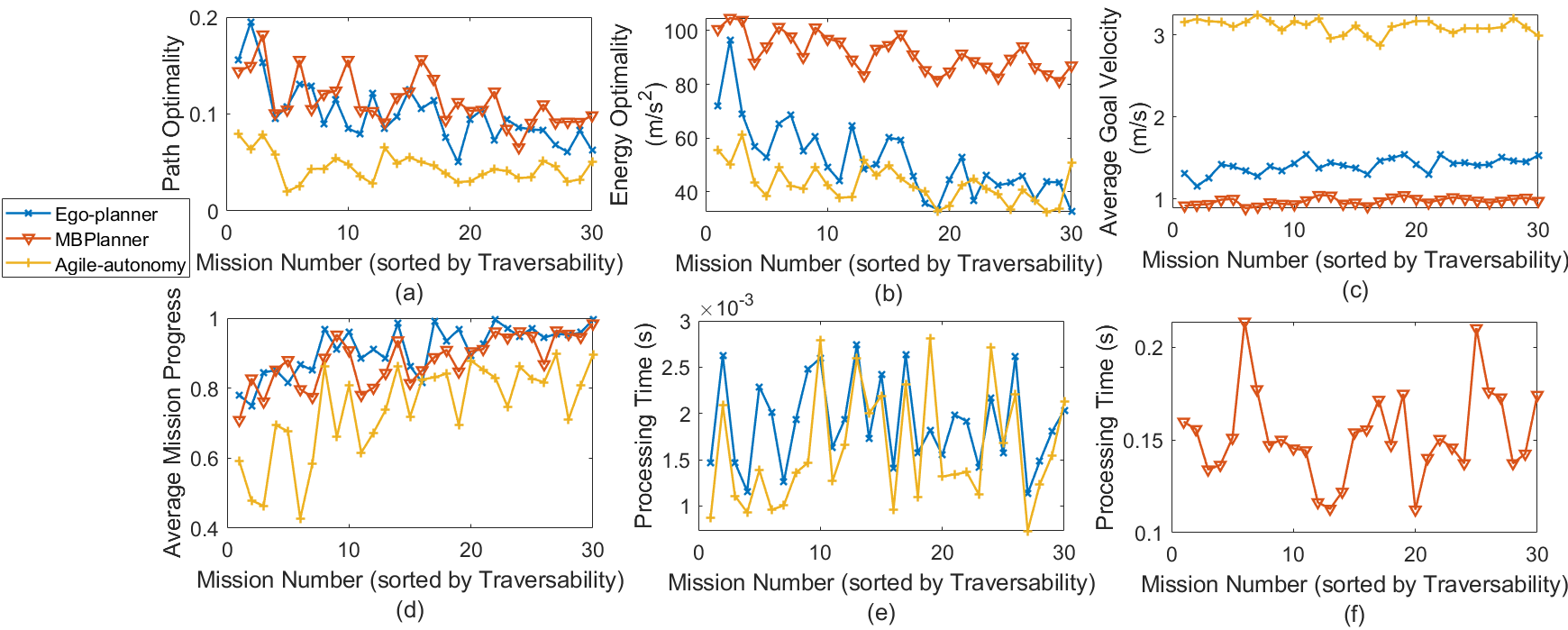}
    \caption{Results of different metrics relating to the increasing
traversability missions. (a) shows the results of path optimality. (b) shows the results of path energy. (c) shows the results of average goal velocity. (d) shows the results of mission progress. (e)-(f) show the results of processing time.}
      \label{figure7}
   \end{figure*}

Figure \ref{figure7}(a) and \ref{figure7}(b) show the result of path optimality and energy optimality respectively for both lower is better. Mission number has been sorted by traversability. The lower of path optimality means drone's trajectory is closer to the minimum distance, and the lower of energy optimality means the drone cost less energy for a same trial. Basically, both Ego-planner and MBPlanner show a downtrend at the metrics of path optimality and energy optimality following the increase of traversability. Moreover, we can see Agile-Autonomy has the best performance in this metric, with a less clear relation to traversability.

Another metric with significant differences is average goal velocity (Figure \ref{figure7}(c)). Agile-Autonomy has been designed for high-speed flight, which results in higher flight speeds than those of Ego-planner and MBPlanner. Mission progress can also indicate the safety of obstacle avoidance. As shown in Figure \ref{figure7}(d), in the results of mission progress, Ego-planner is still the most safe algorithm and we can see mission progress also has a increase trend with traversability.

For the result of processing time(Figure \ref{figure7}(e) and 6(f)), Ego-planner and Agile-autonomy both have the best performance. Although Ego-planner has a process of mapping, it avoids the construction of the ESDF map, thus greatly reducing the processing time. Agile-Autonomy uses an end-to-end neural network, it does not need to build a map, so the running speed is also fast. The redundant map construction of MBPlanner makes its processing time the longest among the three algorithms.

\subsection{Real World Tests}

\subsubsection{Experiment Platform}
\begin{figure}[thpb]
      \centering
        {\includegraphics[width=3.0in]{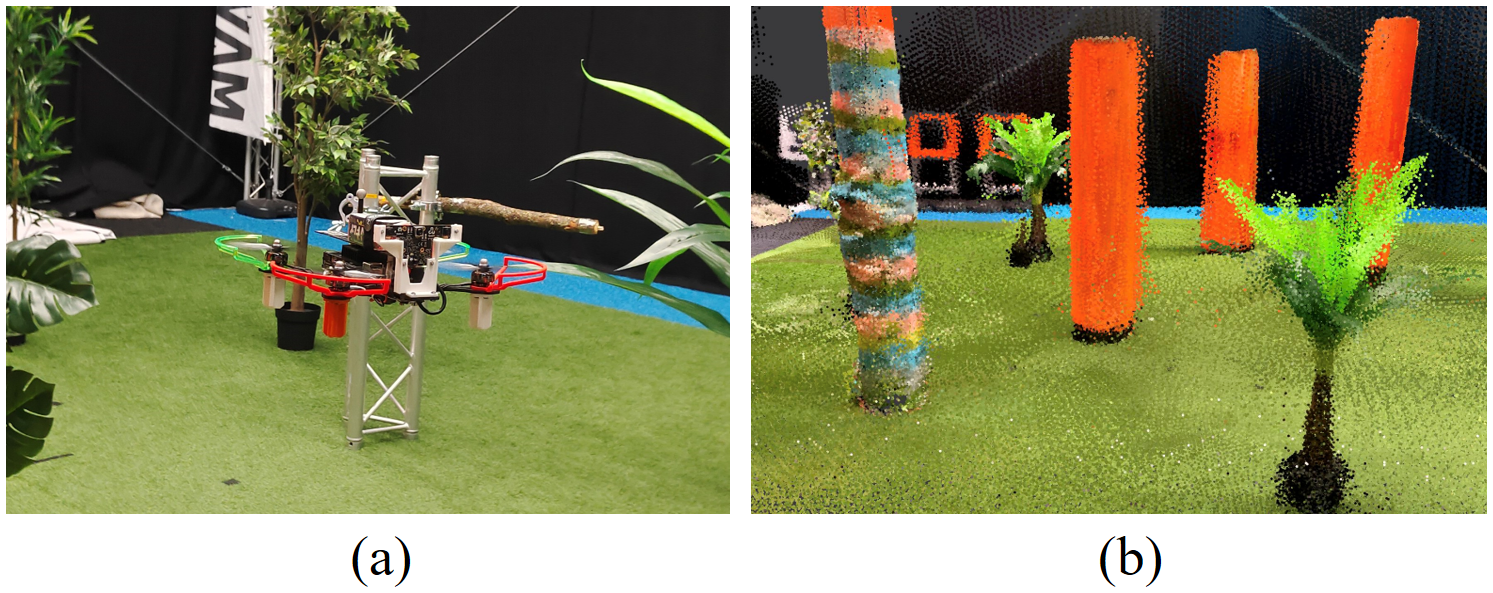}}
    \caption{(a): Experiment platform for real world tests. (b): High-precision point cloud map built by a lidar sensor.}
      \label{figure9}
   \end{figure}

To verify that our proposed metrics can objectively evaluate different algorithms in the real world, we also built a real drone for experimental verification as shown in Figure \ref{figure9}(a). A 6inch frame from Armattan is used for the drone which equipped with Emax 2306 motors and 5inch, three-bladed propellers.

The platform's computational unit is an NVIDIA Jetson Xavier NX module, with 384-core GPU, 48 Tensor Cores, and 6-core ARM CPU. The output of this framework is a low-level control command including ratio of collective thrust and bodyrates to be achieved for flying. The desired commands are sent to a flight controller named Pixhawk mini 4 which running PX4. The drone is also equipped with a stereo camera, Oak-D which can get a dense depth in 400p resolution at 30 Hz.
   
\subsubsection{Experiment Results}

    \begin{figure}[thpb]
      \centering
      {\includegraphics[width=3.4in]{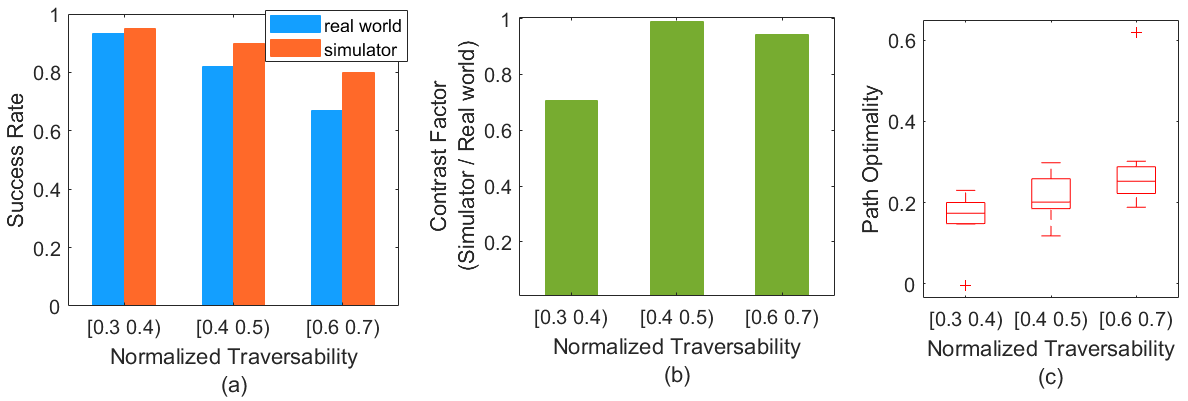}}
      \caption{Comparison of real world tests and simulation tests. (a) shows the results of success rate when real world maps have similar traversability with simulator. (b) shows the results of contrast factor of "simulator / real world". (c) shows the results of path optimality in real world tests.}
      \label{figure14}
   \end{figure}
   
For real world experiments, we arrange three maps of different complexity. To calculate the traversability of real world, we use a lidar sensor to build high-precision point cloud maps, one of the map is shown as Figure \ref{figure9}(b). And for each map, we run 20 trials. From (\ref{eq12}), the normalized traversability can be obtained, then we find the experimental results of simulation in the corresponding interval according to the normalized traversability of real world, and the results are shown as Figure \ref{figure14}.

From Figure \ref{figure14}(a), we can see at similar normalized traversability, simulation results are always better than those of real world. And we also calculate the contrast factor of "Simulation / Real world" as shown in Figure \ref{figure14}(b). From this metric, the influence of distance can be removed, so the results are closed to 1 except the easiest map (This may be caused by the limitation of the number of experiments). And we also show the results of path optimality as shown in Figure \ref{figure14}(c), obviously, this metric has a increasing trend with traversability as we mentioned in simulation part.

\section{CONCLUSIONS}
In this work, we proposed AvoidBench which can evaluate the performance of obstacle avoidance algorithms. In AvoidBench, we can easily set many maps with different obstacles. We combined both performance and environment metrics to explore the performance of different algorithms with increasing environment complexity. Both of the simulation and real world experiments indicate our proposed metrics are reasonable. And the comparison with the real world experiment data also shows that AvoidBench can realize high-fidelity simulation experiments.

\addtolength{\textheight}{-12cm}   % This command serves to balance the column lengths

%%%%%%%%%%%%%%%%%%%%%%%%%%%%%%%%%%%%%%%%%%%%%%%%%%%%%%%%%%%%%%%%%%%%%%%%%%%%%%%%

%%%%%%%%%%%%%%%%%%%%%%%%%%%%%%%%%%%%%%%%%%%%%%%%%%%%%%%%%%%%%%%%%%%%%%%%%%%%%%%%

\bibliographystyle{IEEEtran}
\bibliography{main}

\end{document}